\documentclass[journal=jacsat,manuscript=article]{achemso}
\usepackage[version=3]{mhchem} 

\usepackage{kotex}
\usepackage{multirow}
\usepackage{adjustbox}
\usepackage{color}
\usepackage{xcolor}
\usepackage{hyperref}
\usepackage{verbatim}
\usepackage{graphicx}
\usepackage{listings}
\usepackage{color}
\usepackage{amsmath}
\usepackage{amsfonts}
\usepackage{amssymb}

\definecolor{dkgreen}{rgb}{0,0.6,0}
\definecolor{gray}{rgb}{0.5,0.5,0.5}
\definecolor{mauve}{rgb}{0.58,0,0.82}

\author{Jeonghyeon Kim}
\affiliation[Galux]
{Galux Inc, Gwanak-gu, Seoul 08738, Republic of Korea}
\alsoaffiliation[SNU-CHEM]
{Department of Chemistry, Seoul National University, Seoul 08826, Republic of Korea}

\author{Juno Nam}
\affiliation[Galux]
{Galux Inc, Gwanak-gu, Seoul 08738, Republic of Korea}
\alsoaffiliation[SNU-CHEM]
{Department of Chemistry, Seoul National University, Seoul 08826, Republic of Korea}
\author{Seongok Ryu}
\affiliation[Galux]
{Galux Inc, Gwanak-gu, Seoul 08738, Republic of Korea}
\email{seongokryu@galux.co.kr}

\title[An \textsf{achemso} demo]
  {Understanding active learning of molecular docking and its applications}

\abbreviations{IR,NMR,UV}

\begin{document}

\begin{tocentry}

Some journals require a graphical entry for the Table of Contents.
This should be laid out ``print ready'' so that the sizing of the
text is correct.

Inside the \texttt{tocentry} environment, the font used is Helvetica
8\,pt, as required by \emph{Journal of the American Chemical
Society}.

The surrounding frame is 9\,cm by 3.5\,cm, which is the maximum
permitted for  \emph{Journal of the American Chemical Society}
graphical table of content entries. The box will not resize if the
content is too big: instead it will overflow the edge of the box.

This box and the associated title will always be printed on a
separate page at the end of the document.

\end{tocentry}

\begin{abstract}
With the advancing capabilities of computational methodologies and resources, ultra-large-scale virtual screening via molecular docking has emerged as a prominent strategy for in silico hit discovery.
Given the exhaustive nature of ultra-large-scale virtual screening, active learning methodologies have garnered attention as a means to mitigate computational cost through iterative small-scale docking and machine learning model training. 
While the efficacy of active learning methodologies has been empirically validated in extant literature, a critical investigation remains in how surrogate models can predict docking score without considering three-dimensional structural features, such as receptor conformation and binding poses.
In this paper, we thus investigate how active learning methodologies effectively predict docking scores using only 2D structures and under what circumstances they may work particularly well through benchmark studies encompassing six receptor targets.
Initially, we observe that surrogate model-based score-rankings exhibit concordance primarily among samples possessing high docking scores. 
Furthermore, we confirm that top-scored compounds demonstrate substantial three-dimensional shape similarities, where similar structural patterns are related to shape and interaction patterns specific to binding pockets.
To assess the impact of memorization versus generalization of structural and interaction features on predictions, we employ linear factor analysis on the predicted docking score by using molecular descriptors.
Our findings suggest that surrogate models tend to memorize structural patterns prevalent in high docking scored compounds obtained during acquisition steps.
Despite this tendency, surrogate models demonstrate utility in virtual screening, as exemplified in the identification of actives from DUD-E dataset and high docking-scored compounds from EnamineReal library, a significantly larger set than the initial screening pool.
Our comprehensive analysis underscores the reliability and potential applicability of active learning methodologies in virtual screening campaigns.
\end{abstract}

\section{Introduction}

With recent advancements in computational capabilities and the widespread accessibility of ultra-large-scale libraries containing readily purchasable compounds, high-throughput virtual screening (HTVS) via molecular docking has become prevalent in the realm of in silico hit discovery.\cite{lyu2019ultra, gorgulla2020open, bender2021practical, ton2020rapid, mcgann2021gigadockingtm, stein2020virtual} 
Recent works on ultra-large-scale HTVS have demonstrated a direct correlation between the size of the screening library and the attainable hit-finding rate, a concept often referred to as ``the bigger, the better.''
For instance, \citeauthor{lyu2019ultra} screened nearly 100 million compounds from the EnamineREAL library\cite{EnamineREAL} to prioritize active compounds against AmpC and D\textsubscript{4} dopamine target receptors. 
Their efforts yielded a 24\% hit rate among 124 experimental candidates, which were selected by top docking-scored compounds with visually favorable binding modes.\cite{lyu2019ultra}
Additionally, \citeauthor{gorgulla2020open} conducted screening of 1.38 billion compounds and empirically validated that the docking scores of the top 50 compounds exhibit logarithmic increases with the size of the screening library.\cite{gorgulla2020open}
Despite its efficacy, the utilization of ultra-large virtual screening incurs substantial computational costs, particularly as the size of the screening library expands.
For instance, conducting screening on 1.3 billion compounds utilizing 8,000 CPUs requires a running time of 28 days.\cite{gorgulla2020open} Based on standard pricing observed on the Google Cloud Platform, the associated cost is estimated to be at least \$200k.\cite{gcp}

To circumvent exhaustive screening of entire library compounds, researchers have turned to the adoption of active learning\cite{cohn1996active} and Bayesian optimization\cite{snoek2012practical} frameworks, which facilitate efficient exploration of chemical space guided by surrogate predictors.\cite{gentile2020deep, gentile2022artificial, graff2021accelerating, yang2021efficient, marin2023regression, cavasotto2023impact, chen2024active} 
This framework is designed to identify the highest docking-scored compounds, within resource constraints, through iterative running of molecular docking and machine learning model.
In brief, a machine learning model is trained to fit docking scores, which are computed in preceding simulation steps, using structural descriptors.
Subsequently, the trained model predicts and selects compounds most likely to yield high docking scores. 
Docking simulations are then conducted on the selected samples, and the resulting data are used to further refine the machine learning model in subsequent training iterations. 
Through this iterative process of docking, training, and inference with an appropriate acquisition strategy, previous benchmarks have demonstrated the discovery of top-docking-scored compounds with a success rate exceeding 90\%, while requiring less than 10\% of the simulation time required for docking the entire library.\cite{graff2021accelerating}

Numerous studies have attempted to enhance the efficiency and efficacy of active learning protocols.
\citeauthor{graff2022self}\cite{graff2022self} successfully reduced computational costs without compromising performance by pruning poor-performing candidates. 
\citeauthor{bellamy2022batched}\cite{bellamy2022batched} examined diverse acquisition methods within noisy setups, affirming the robustness of the greedy acquisition approach under such conditions. 
Additionally, \citeauthor{pyzer2018bayesian}\cite{pyzer2018bayesian} proposed a framework for selecting appropriate acquisition methods tailored to specific circumstances.
While simple acquisition strategies, like greedy acquisition, demonstrate effectiveness in simple datasets, more sophisticated approaches leveraging predictive uncertainty, such as expected improvement (EI) acquisition, are advocated for challenging datasets.
This study may elucidate the widespread adoption of greedy methods in prior endeavors employing active learning with molecular docking. 

In addition to the empirical success of active learning methodologies, we underscore the significance of elucidating the mechanisms underlying their operation.
Active learning of molecular docking diverges from conventional structure-based virtual screening, as surrogate models often predict docking scores devoid of receptor information, relying solely on ligand structural features.
However, intriguing findings indicating the accurate prediction of docking scores solely through 2D information of ligands prompt a reevaluation of the necessity of computationally intensive three-dimensional information. 
In light of these empirical observations, we are compelled to question whether surrogate models merely memorize common patterns within top-scored compounds or generalize underlying principles governing protein-ligand interactions.
This inquiry extends beyond molecular docking to encompass the learnability of diverse molecular properties, such as cell viability (bio-activity), toxicity, and metabolism, absent three-dimensional structural insights.

This study delves into the intricacies of active learning within molecular docking processes, employing six receptor targets and compounds pools sourced from widely utilized screening libraries, namely EnamineHTS\cite{EnamineHTS} and EnamineREAL\cite{EnamineREAL}, alonside actives and decoys extracted from the DUD-E\cite{mysinger2012directory} datasets.
Our analysis confirms that the utilization of acquisition strategies tailored to sample top-scored compounds, such as Greedy and upper confidence bound (UCB) acquisitions, engenders a biased exploration of  chemical space where top-scored compounds are populated.
This bias arises from machine learning models memorizing common structural patterns prevalent in top-scored samples, likely originating from shared shape and interaction patterns specific to binding pockets.
Despite the dependency of surrogate models on memorization, our findings demonstrate that their applicability for screening across various molecular pools, exemplified by the successful identification of actives from the DUD-E dataset and high docking-scored compounds sourced from the 100M sized EnamineREAL library.

\section{Methods}

\subsection{Workflow of active learning}
\begin{figure}[hbt!]
    \centering
    \includegraphics[width=15cm, height=15cm, keepaspectratio]{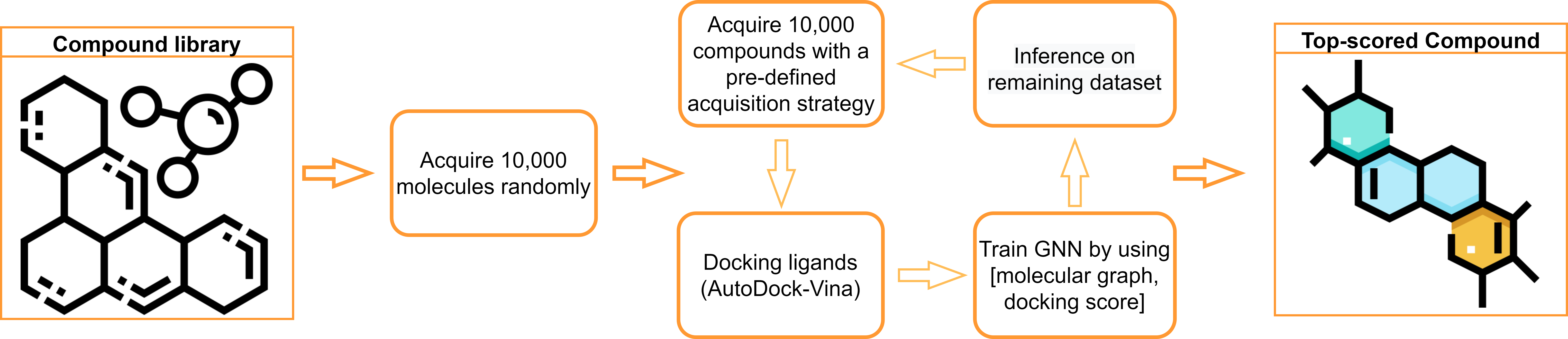}
    \caption{Our workflow for active learning of molecular docking. We used the EnamineHTS\cite{EnamineHTS} library as a pool of ligands and AutoDock Vina\cite{trott2010autodock} as a docking simulation tool.}
    \label{fig:AL_protocol}
\end{figure}

Active learning or Bayesian optimization framework aims to identify samples most likely to fulfill pre-defined objectives.
These objectives typically fall into two primary categories: obtaining a maximal number of high-scored samples and enhancing the predictive performance of the model to the greatest extent possible. 
To achieve these aims, the framework iteratively conducts experiments or simulations to acquire labeled samples and subsequently trains a surrogate model with the newly acquired samples. 

In this study, our objectives are to collect samples exhibiting the highest docking scores and to optimize the predictive accuracy of the surrogate model.
While these objectives are commonly known as Bayesian optimization and active learning, respectively, we henceforth refer to our approach as active learning, irrespective of the acquisition strategy employed.
The overall active learning workflow implemented in this study is depicted in Figure \ref{fig:AL_protocol}.
Initially, 10,000 ligands were randomly sampled from the comprehensive EnamineHTS library\cite{EnamineHTS} and subjected to docking simulations with a desired receptor. 
Subsequently, a graph neural network\cite{battaglia2018relational} was trained to estimate the predictive docking score and its heteroscedastic aleatoric uncertainty\cite{kendall2017uncertainties} based on input molecular graphs. 
Upon completion of the training phase, an additional 10,000 ligand compounds were sampled from the remaining screening library utilizing three acquisition strategies: Greedy, Upper Confidence Bound (UCB), and Uncertainty (UNC).
\begin{equation}
\begin{split}
    \text{Greedy:}& \quad a(x) = \hat{y}(x)\\
    \text{UCB:}& \quad a(x) = \hat{y}(x) + 2\hat{\sigma}(x)\\
    \text{UNC:}& \quad a(x) = \hat{\sigma}(x),
\end{split}
\end{equation}
where $\hat{y}(x)$ and $\hat{\sigma}(x)$ represent the predictive mean and uncertainty, respectively, of docking score corresponding to the given molecular structure $x$.
We note that the adoption of Greedy or UCB acquisition strategies is geared towards the collection of samples anticipated to exhibit the highest docking scores, aligning with the principles of Bayesian optimization. 
Conversely, UNC acquisition is tailored to gather samples characterized by the highest expected uncertainty, thereby enhancing prediction accuracy across the entire spectrum of docking scores. 
Employing each of these strategies, we trained the GNN model over ten acquisition steps, encompassing an initial random acquisition followed by nine strategic acquisitions. 

\subsection{Dataset preparation}

\begin{table}[]
\caption{List of targets used in our benchmark test. The group, docking quality, and regression $R^2$ values are as reported in DOCKSTRING\cite{garcia2022dockstring}, except for SOS1.}
\label{tab:targets}
\begin{tabular}{|c|c|c|c|c|}
\hline
       & PDB ID & Group            & Docking Quality & Regression $R^2$ \\ \hline
EGFR   & 2RGP   & Kinase           & **              & -             \\ \hline
JAK2   & 3LPB   & Kinase           & ***             & 0.853         \\ \hline
PGR    & 3KBA   & Nuclear receptor & ***             & 0.678         \\ \hline
ESR2   & 2FSZ   & Nuclear receptor & ***             & 0.627         \\ \hline
PARP1  & 3L3M   & Enzyme           & ***             & 0.910         \\ \hline
SOS1   & 6SCM   & Enzyme           & -               & -             \\ \hline
\end{tabular}
\end{table}

We used the EnamineHTS library\cite{EnamineHTS}, which comprises 2.1M molecules, as the pool of ligands for conducting docking simulations across six receptor targets detailed in Table \ref{tab:targets}.
Following an assessment of benchmark regression results on the DUD-E dataset\cite{mysinger2012directory} within the DOCKSTRING\cite{garcia2022dockstring}, 
we identified two high regression $R^2$ targets (JAK2 and PARP1), two low $R^2$ targets (PGR and ESR2), and one unknown $R^2$ target (EGFR) for inclusion in our benchmarking analysis.
Additionally, we incorporated SOS1 as a target system of interest, actively explored in recent drug discovery campaigns\cite{ketcham2022design, ramharter2021one}. 

\section{Results and Discussion}
\subsection{Benchmark results}

For the benchmark study, we randomly partitioned two sets of 10,000 compounds from the 2.1M compounds within the EnamineHTS library to form the validation and test sets, enabling the monitoring of training progress and the assessment of prediction performance. 
Herein, we report the mean and standard deviation of test set results obtained across four runs employing distinct random seeds. 

\begin{figure}[hbt!]
    \centering
    \includegraphics[width=0.95\textwidth,trim={0cm 0 0cm 0},clip]{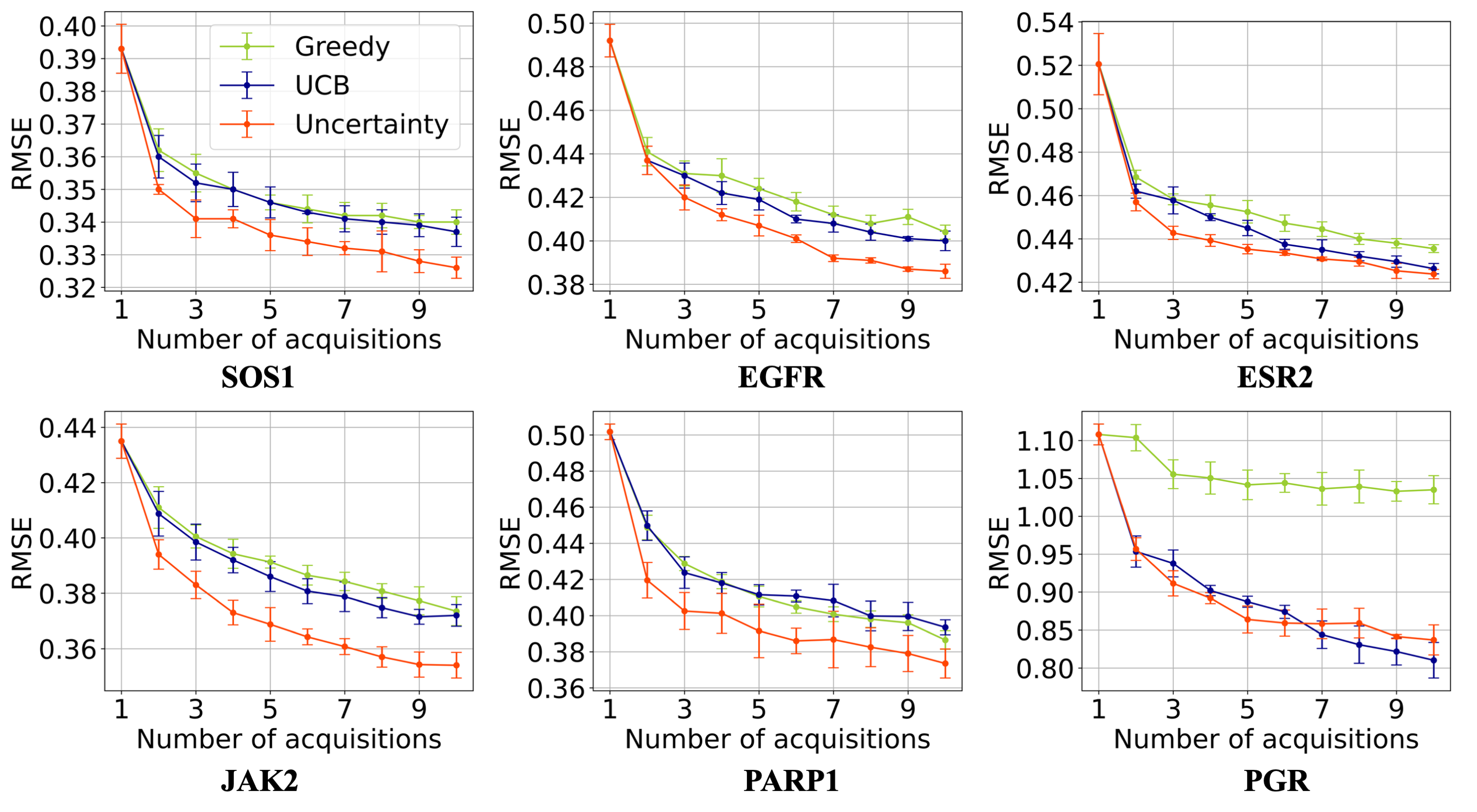}
    \caption{The root-mean-square error (RMSE) between docking scores and prediction scores plotted against the number of acquisitions, with varying acquisition strategies and for six receptor targets.}
    \label{fig:RMSE_gine}
\end{figure}

\begin{figure}[hbt!]
    \centering
    \includegraphics[width=0.95\textwidth,trim={0cm 0 0cm 0},clip]{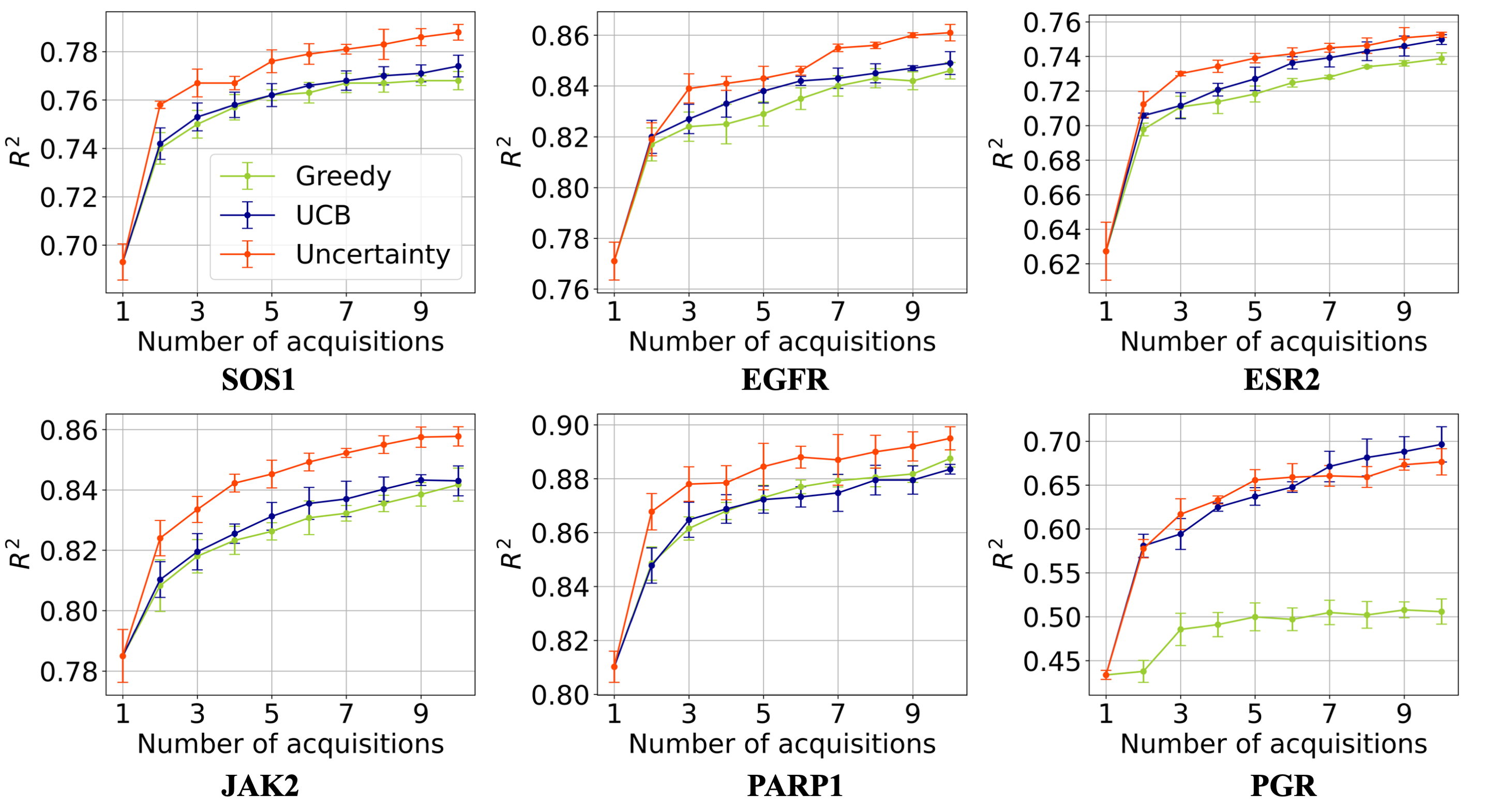}
    \caption{The coefficient of determination $R^{2}$ between docking scores and prediction scores plotted against the number of acquisitions.}
    \label{fig:R^2_gine}
\end{figure}

\begin{figure}[hbt!]
    \centering
    \includegraphics[width=0.95\textwidth,trim={0cm 0 0cm 0},clip]{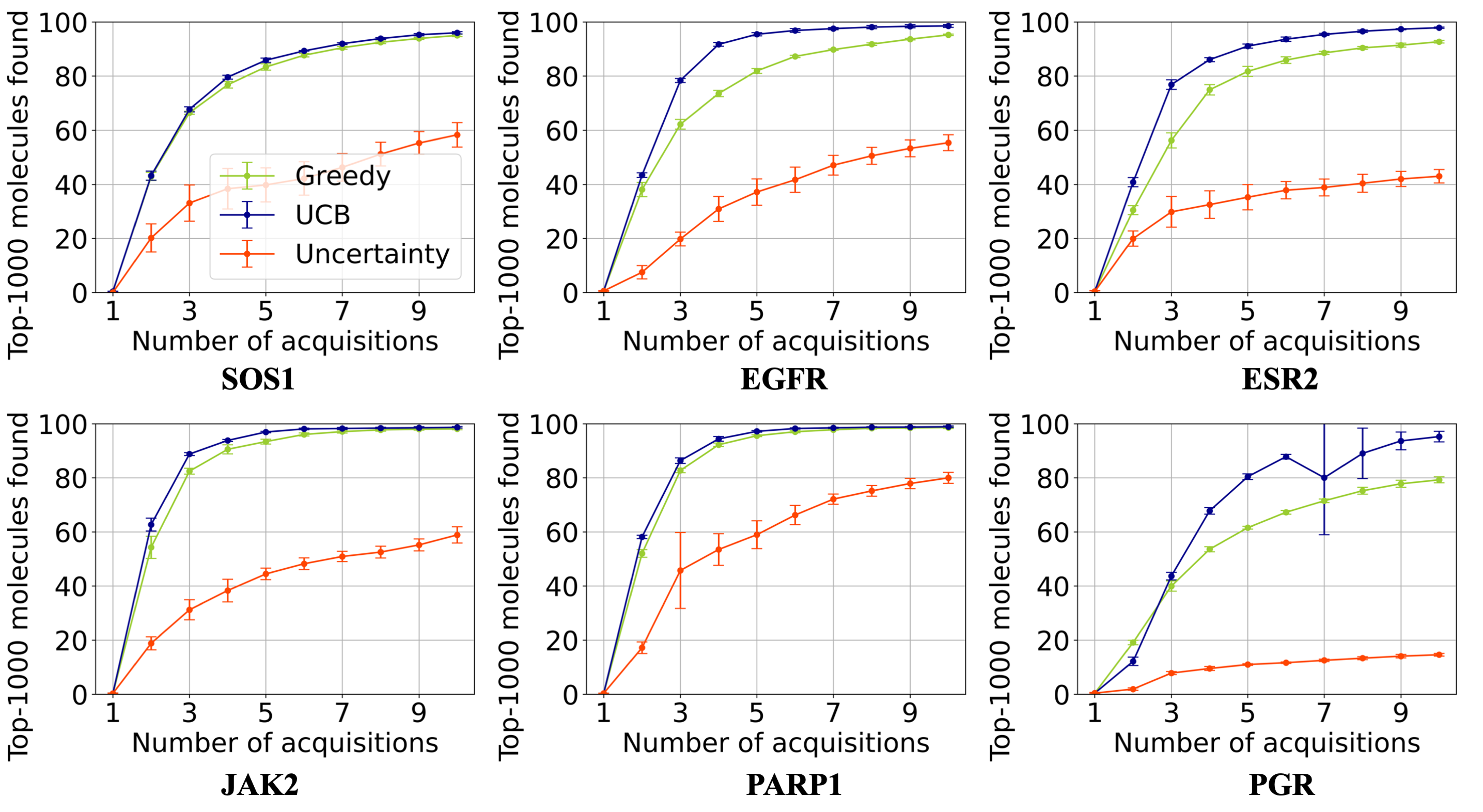}
    \caption{The recovery rate of top-1000 compounds plotted against the number of acquisitions.}
    \label{fig:recovery_gine}
\end{figure}

Figures \ref{fig:RMSE_gine}, \ref{fig:R^2_gine}, and \ref{fig:recovery_gine} track the evolution of prediction accuracies, measured by the root-mean-square error (RMSE) and coefficient of determination ($R^2$) between the docking and prediction scores, alongside the top-1000 recovery rate.
Notably, with an increase in the number of acquisitions, the prediction accuracy (RMSE and $R^2$) is gradually improved.
The uncertainty acquisition consistently outperforms the greedy and UCB acquisitions in terms of prediction accuracy. 
Across all receptor targets and acquisition strategies, the top-1000 recovery rates exhibit a steady increase, nearly reaching saturation at the final acquisition step. 
As anticipated prior to experimentation, both Greedy and UCB acquisitions demonstrate superior top-1000 recovery rates compared to uncertainty acquisition, as they are specifically designed to acquire samples expected to possess high docking scores.
Conversely, uncertainty acquisition excels in prediction accuracy (RMSE and $R^2$), given its focus on identifying the most probable samples to improve model performance.
Therefore, if the objective is to maximize the discovery of top-docking-scored compounds within limited computational resources, employing Greedy or UCB acquisition strategies is recommended.

Furthermore, we observe that the predictive performance is contingent upon the predictability of the docking score, as outlined in Table \ref{tab:targets}.
Upon completion of the iterative acquisitions, the JAK2 and PARP1 receptors, with DOCKSTRING\cite{garcia2022dockstring} regression $R^2$ values of 0.853 and 0.910, exhibit lower RMSE and higher prediction $R^2$ compared to the ESR2 and PGR receptors, with DOCKSTRING regression $R^2$ values of 0.627 and 0.678, respectively.
Particularly noteworthy is the notably inaccurate prediction performance of the PGR receptor relative to the other five receptors, accompanied by lower and slower convergence of recovery rates as the number of acquisition steps increases.

\subsection{Successful recovery of top-docking-scored samples was observed only for top-prediction-scored samples}

\begin{figure}[hbt!]
    \centering
    \includegraphics[width=0.9\textwidth,trim={0cm 0 0cm 0},clip]{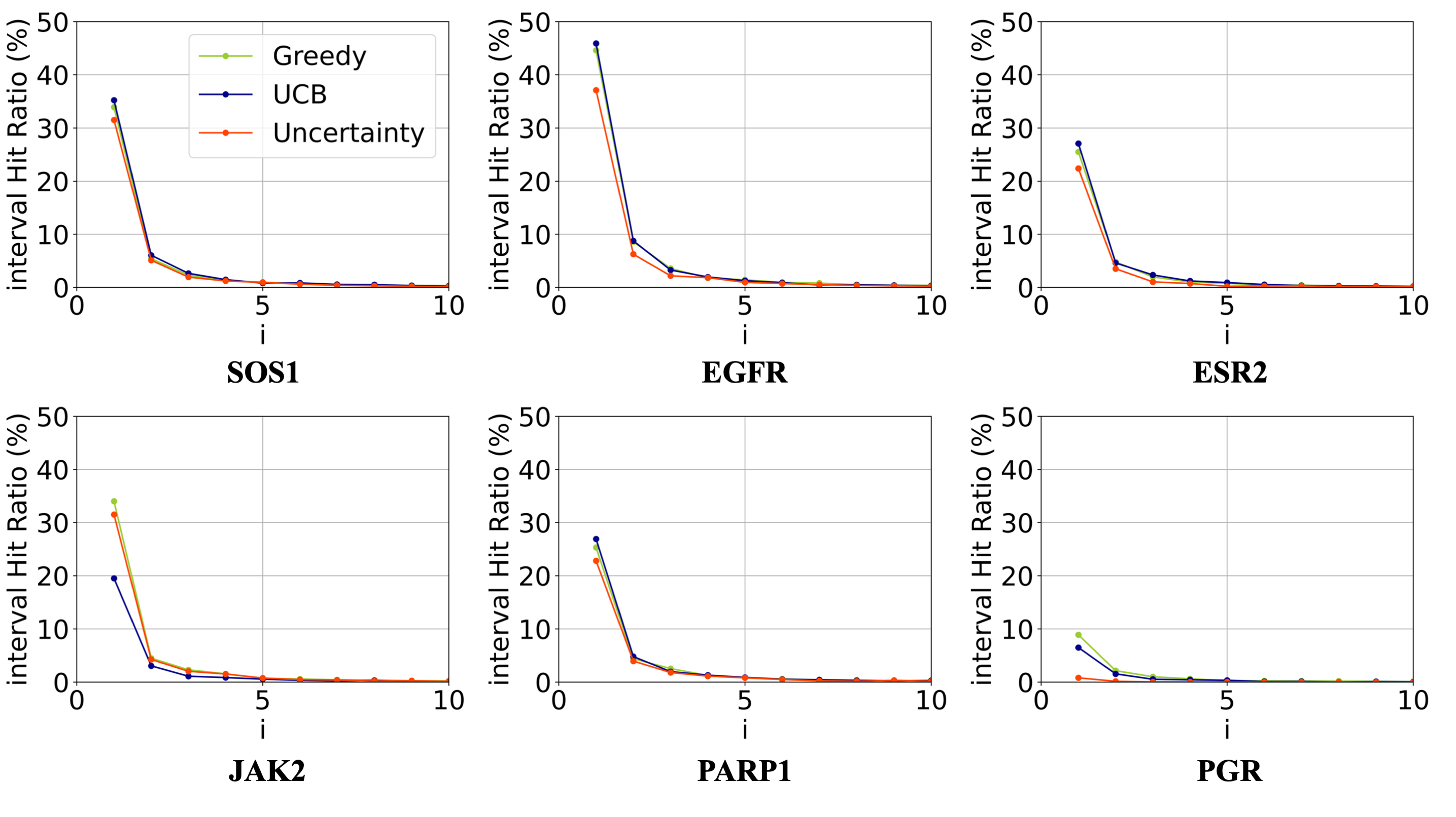}
    \caption{Change of $i$-th interval hit ratio (eq~\eqref{eq:HR}) as the interval index $i$ increases. Predictions from the surrogate GNN models are accurate only for top-docking-scored compounds.}
    \label{fig:HR_i,k}
\end{figure}

\begin{figure}[hbt!]
    \centering
    \includegraphics[width=0.9\textwidth,trim={0cm 0 0cm 0},clip]{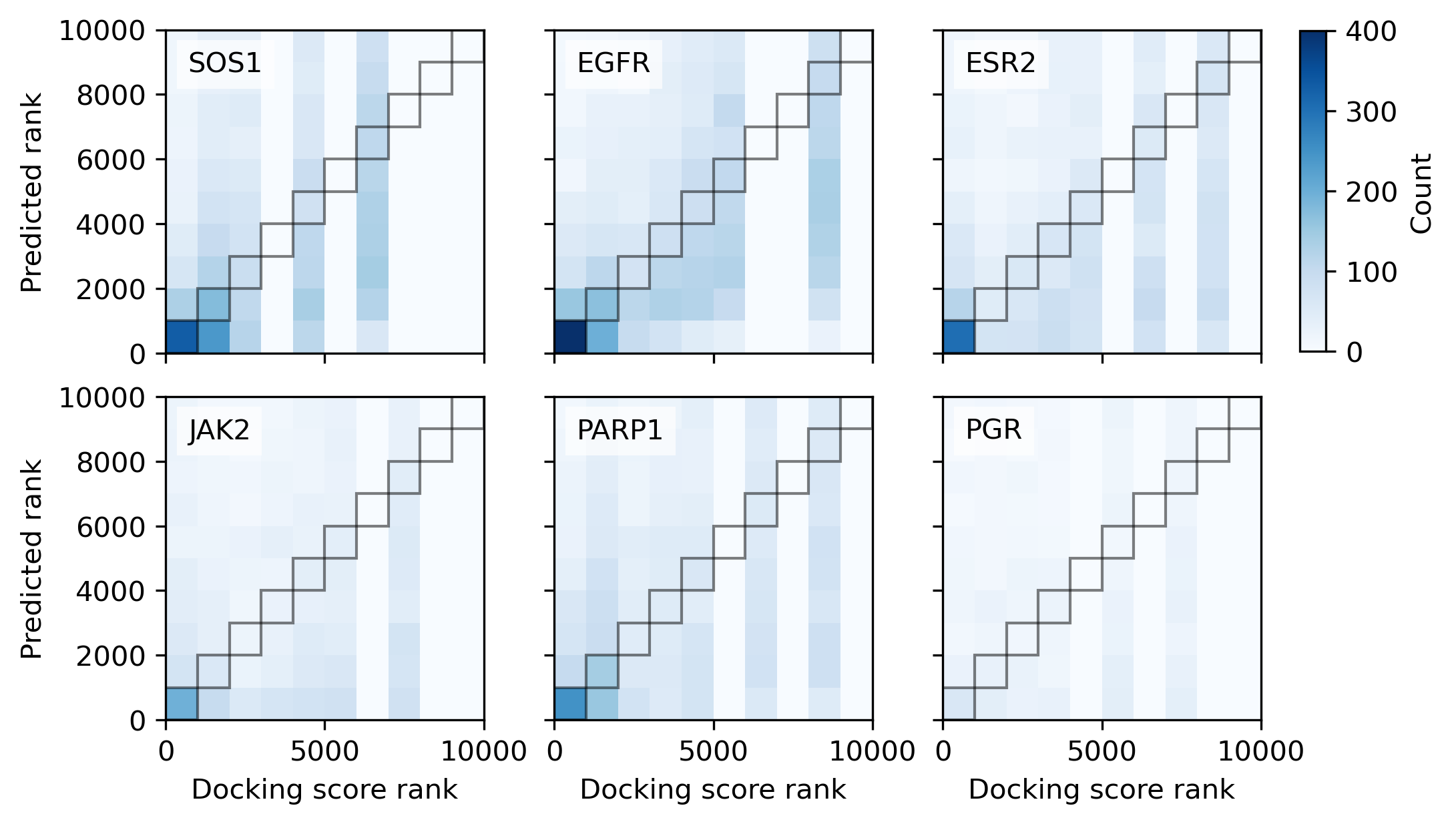}
    \caption{2-D histogram of rank order between docking and prediction scores obtained by the UCB acquisition. Predictions from the surrogate GNN models were accurate only for top-docking-scored compounds.}
    \label{fig:Hexbin plot}
\end{figure}

Given the biased sampling induced by Greedy and UCB acquisitions across the entire pool, we scrutinized the validity of surrogate predictors' prediction of docking scores and the recovery of top-scored compounds for the entire set, as opposed to solely those with top prediction scores.
To evaluate the discovery rate within a designated docking score rank interval, we define $i$-th interval hit ratio, $\text{HR}(i,k)$, as
\begin{equation}
    \text{HR} (i,k) = \frac{| \{ (i-1)  \times k < \text{Rank}(\text{Dock}) \leq i \times k \} \cap \{ (i-1) \times k < \text{Rank}(\text{Pred}) \leq i \times k \} |}{k}.
    \label{eq:HR}
\end{equation}
$\text{HR}(i=1,k)$ focuses on the intersection between top-$k$ docking-scored and top-$k$ prediction-scored samples.
Given our interest in assessing rank matching beyond the top-$k$ scored samples, we extend our investigation to $\text{HR}(i,k)$ for $i > 1$.
Accordingly, we set $k=1,000$ and explore $\text{HR}(i,k)$ for $i \in [1,10]$.
Note that the prediction scores were obtained by using the final surrogate predictors trained with 100,000 compounds acquired through ten acquisition steps.

In Figure \ref{fig:HR_i,k}, we illustrate the variation in $\text{HR}(i,k=1,000)$ across different interval indices $i$ for our six receptors. Notably, for all receptors, $\text{HR}(i,k)$ exhibits a rapid decrease as $i$ increases, indicating a diminished accuracy of the surrogate GNN models for higher docking-scored compounds attributed to biased sampling.
In essence, the Greedy and UCB acquisition functions were found to be suitable for identifying high docking-scored samples rather than ensuring a diverse sample selection. 
From the 2-D histogram illustrating the rank order correlation between  docking scores and the prediction scores derived from the UCB surrogate model (Figure~\ref{fig:Hexbin plot}), we reaffirm that prediction results hold validity primarily for top-prediction-scored samples.
The highest number of correctly ordered samples is observed among the top-1000 samples, with a significant decrease in accuracy as the rank order ascends. 
Moreover, a higher prediction score provided by the surrogate model corresponds to a more accurate rank order of docking score.
Consequently, while exploring a larger number of samples may lead to the discovery of more hit samples, the recovery rate is expected to gradually decline.

\subsection{Characteristics of samples acquired by surrogate predictors} 
\begin{figure}[hbt!]
    \centering
    \includegraphics[width=1.0\textwidth,trim={0cm 0 0cm 0},clip]{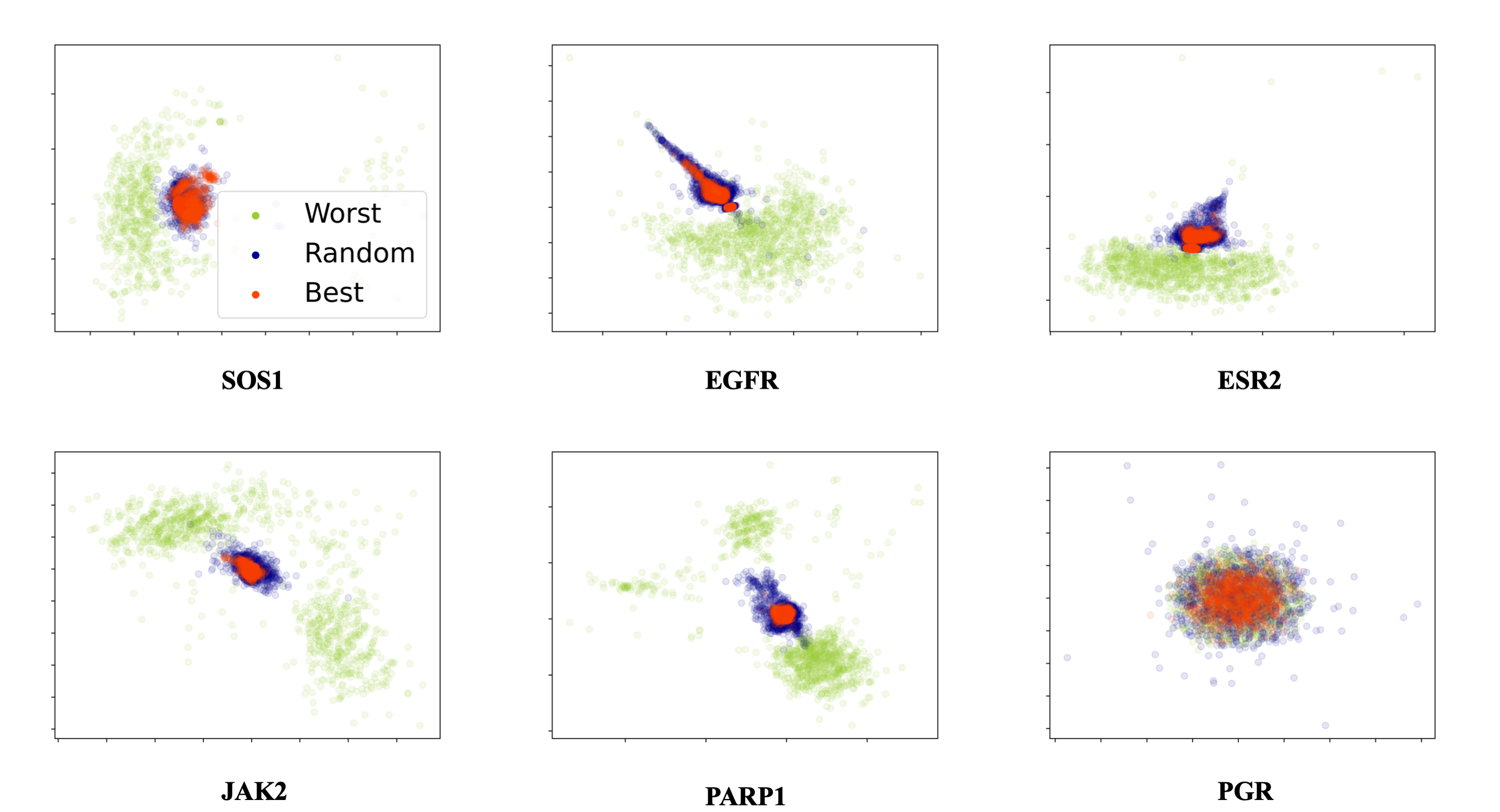}
    \caption{The MDS projection of three-dimensional structure  among the top-1000 compounds (Best), randomly sampled 1000 compounds (Random), and the worst 1000 compounds (Worst) based on their docking scores evaluated on the six receptor targets.}
    \label{fig:3d_similarity}
\end{figure}

Subsequently, we delve into the characteristics of compounds identified by the surrogate models. 
We posit that there are common traits among top-scored compounds, which enables more accurate predictions by the surrogate model.
For instance, compounds with high prediction scores tend to share structural features conducive to a well-suited three-dimensional shape within a given binding pocket. 
In accordance with the principles of structure-activity relationship, compounds exhibiting structural similarity often display comparable binding affinities.

Therefore, to explore whether top-scored compounds demonstrate three-dimensional (3D) structural similarity across binding conformations, we computed the 3D structural distances among compounds within the three distinct groups: (1) the top-1000 prediction-scored compounds, (2) a randomly sampled subset of 1000 compounds, and (3) the bottom 1000 prediction-scored compounds.
In Figure \ref{fig:3d_similarity}, we visualize the distance map using multi-dimensional scaling (MDS)\cite{cox2008multidimensional}.

With the exception of the PGR receptor, we observed a close distribution of samples within the top-prediction-scored groups, consistent with our hypothesis that top-scored compounds share similar 3D structural (binding pose) patterns.
Additionally, the top-prediction-scored groups were spatially separated from both the worst-prediction-scored groups and randomly-sampled groups. 
Therefore, we further hypothesized that the surrogate GNN models might discern structural patterns from the acquired compounds, an assertion we aim to verify in subsequent experiments. 
Conversely, for the PGR receptor, the clustering pattern appeared significantly less discernible compared to other receptors; the MDS visualization points lacked clear spatial separation.
This outcome suggests that the surrogate GNN models encountered difficulty in learning patterms from the acquired compounds, as also observed in Figures \ref{fig:RMSE_gine}, \ref{fig:R^2_gine}, and \ref{fig:recovery_gine}.

\begin{figure}[hbt!]
    \centering
\includegraphics[width=1.0\textwidth,trim={0cm 0 0cm 0},clip]{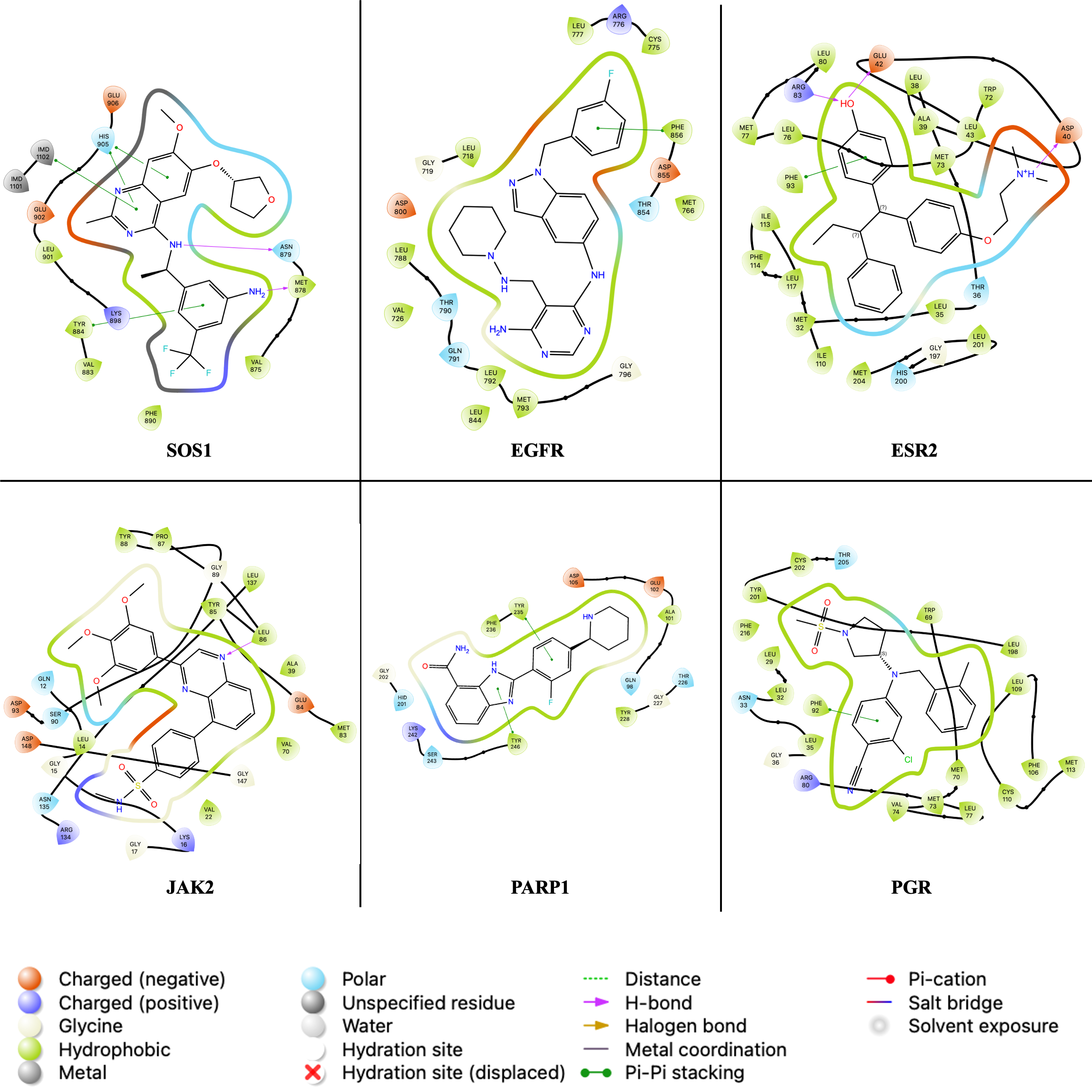}
    \caption{Binding pose between the reference ligand and the receptor for SOS1, EGFR, ESR2, JAK2, PARP1, and PGR complexes.}
    \label{fig:ref_molecule}
\end{figure}

To comprehend the distinct prediction behavior between the PGR receptor and the remaining receptors, we investigated characteristics of the binding pocket and binding pose results for each receptor.
In Figure \ref{fig:ref_molecule}, we show the binding poses alongside key pharmacophore interactions for our six receptors. 
Generally, protein-ligand complexes with high binding affinity manifest well-characterzied ligand fitting to given binding pockets, characterized by restrained flexibility and interactions with side chains or backbones.
Notably, the first five receptors exhibited such characteristics.
Conversely, the binding pocket of the PGR receptor appeared wide and open-shaped, with the majority of pharmacophore interactions being hydrophobic.
We hypothesize that the absence of binding pocket-specific interactions may contribute to a more diverse docking pose, ultimately resulting in the diminished predictability of docking scores.

\begin{figure}[hbt!]
    \centering
    \includegraphics[width=0.8\textwidth,trim={0cm 0 0cm 0},clip]{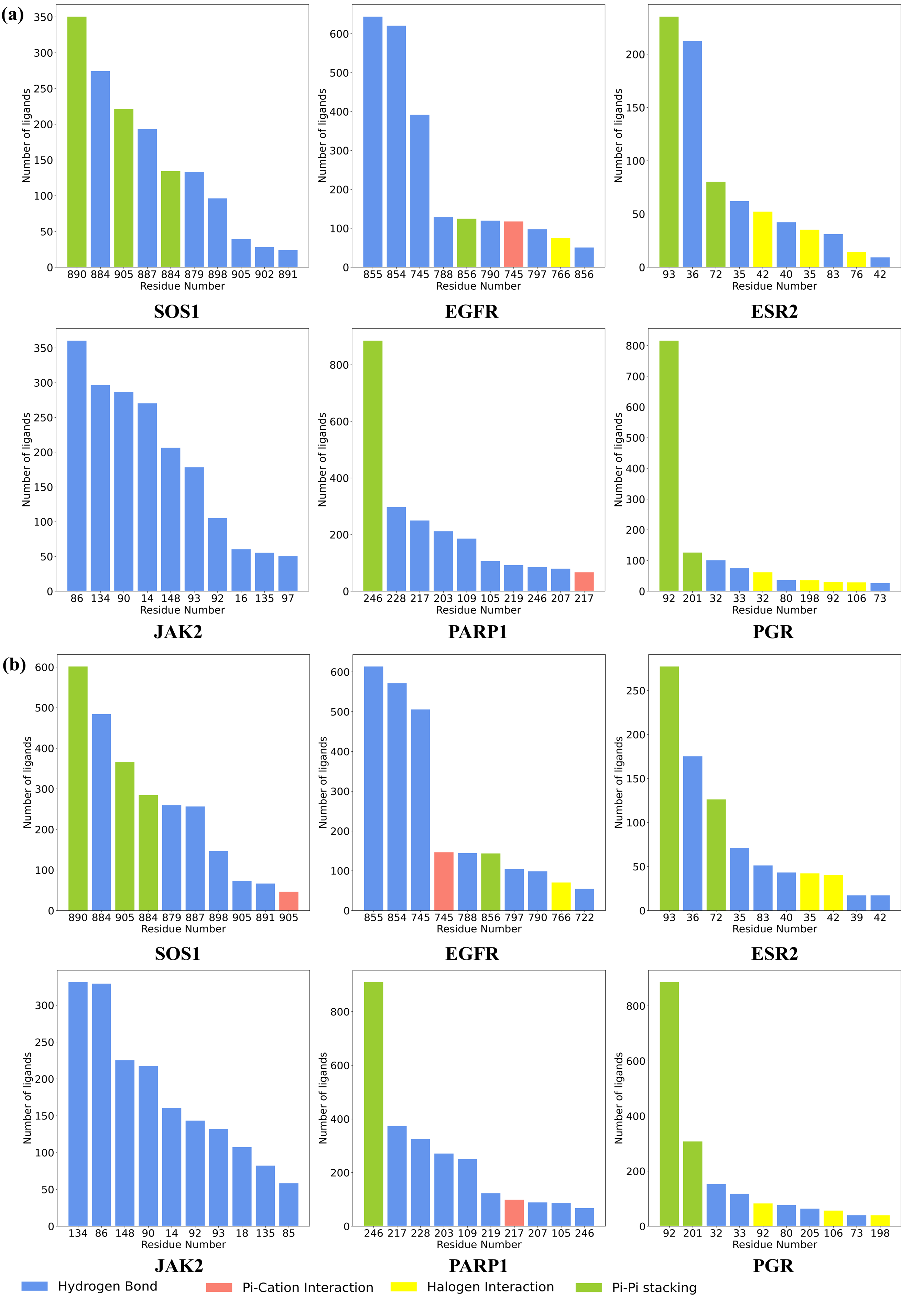}
    \caption{Distribution of pharmacophore interactions in a) top-1000 GNN prediction scored compounds b) top-1000 Docking scored compounds. Note that hydrophobic interactions are excluded in this figure.}
    \label{fig:Pharmacophore}
\end{figure}

\begin{figure}[hbt!]
    \centering
    \includegraphics[width=1.0\textwidth,trim={0cm 0 0cm 0},clip]{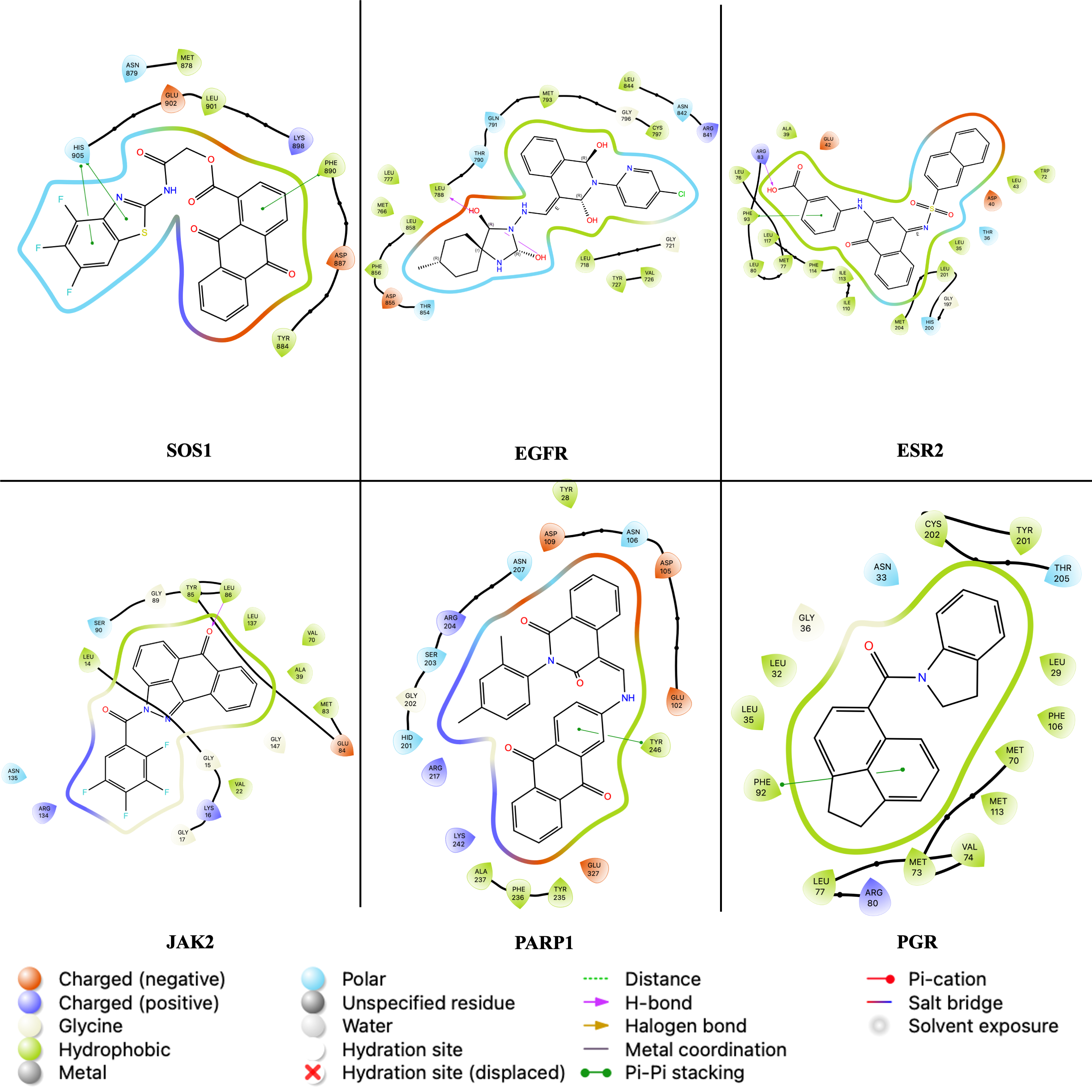}
    \caption{Binding pose of Top-1 GNN prediction scored compounds for visual reference.}
    \label{fig:Pharmacophore_bindingpose}
\end{figure}

\begin{figure}[hbt!]
    \centering
    \includegraphics[width=1.0\textwidth,trim={0cm 0 0cm 0},clip]{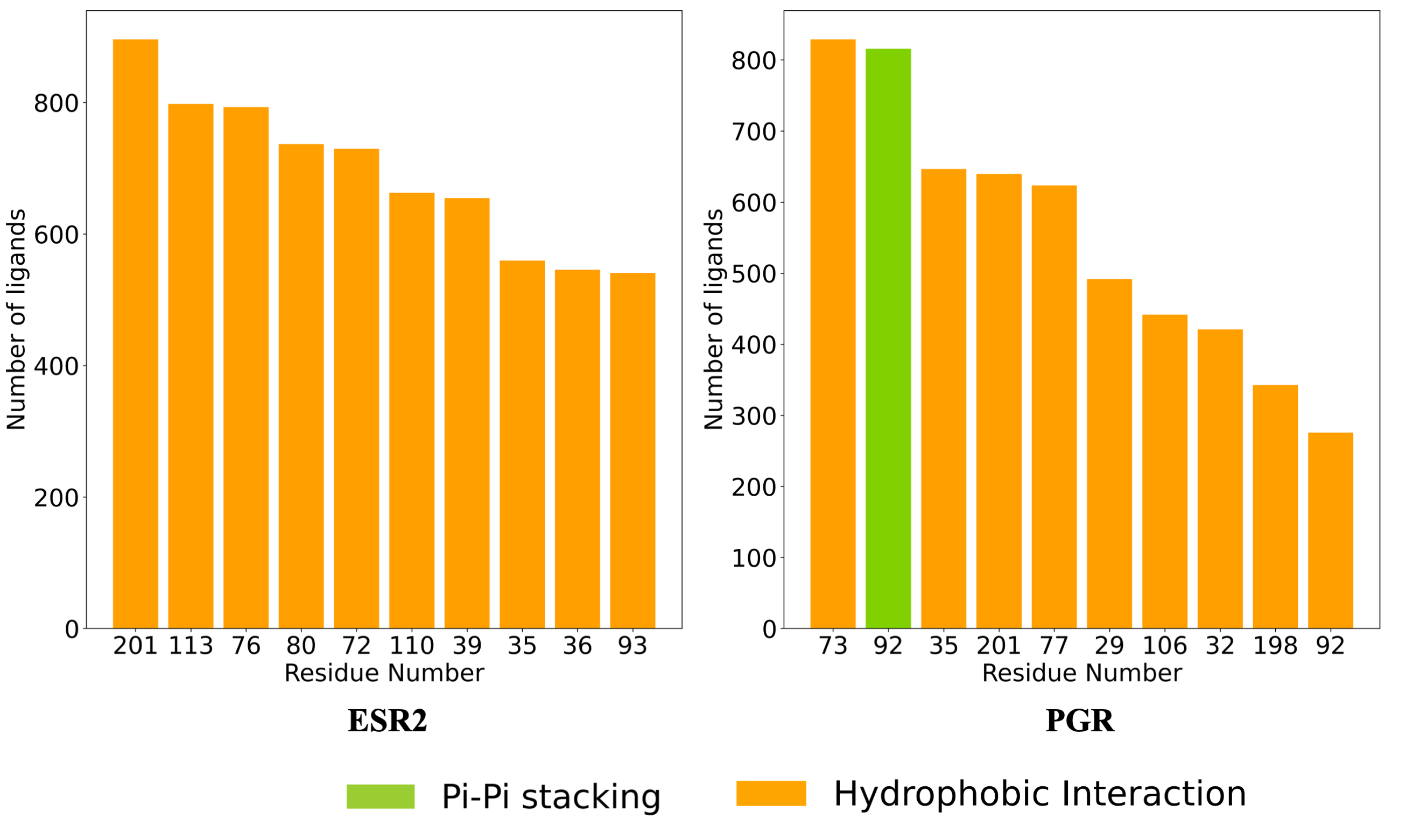}
    \caption{Distribution of pharmacophore interactions in top-1000 GNN prediction scored compounds including hydrophobic interaction for ESR2 and PGR.}
    \label{fig:ESR2_PGR}
\end{figure}

For further analysis, we investigated common pharmacophore interactions observed in the docking poses.
In Figure \ref{fig:Pharmacophore}, we show the top-10 pharmacophore interactions present in both the top-1000 prediction-scored and top-1000 docking-scored compounds. 
Furthermore, in Figure \ref{fig:Pharmacophore_bindingpose}, we provide a visualization of the binding mode of the highest prediction-scored examples for reference.

In the case of SOS1, we noted frequent pi-pi stacking interactions with residues $\text{PHE}^{890}$, $\text{HIS}^{905}$, and $\text{TRY}^{884}$. 
Notably, known SOS1 inhibitors such as BI-3406 (PDB ID: 6SCM) and MRTX-0902 (PDB ID: 7UKR) exhibit pi-pi stacking with residues $\text{PHE}^{890}$ and $\text{HIS}^{905}$.
Additionally, hydrogen bonding interactions with residues $\text{TYR}^{884}$, $\text{ASP}^{887}$, and $\text{ASN}^{879}$ were among the top-10 interactions observed.
We concluded that these interactions were pivotal for ligands to aptly fit within the SOS1 binding pocket, leading to the presence of `Ring -- Bridge -- Ring' structure patterns in top-scored compounds.
Similar pharmacophore interaction patterns were identified in the ESR2 case, wherein pi-pi stacking interactions with residues $\text{PHE}^{93}$ and $\text{TRP}^{72}$  predominated.
Furthermore, hydrogen bonding interactions with residues $\text{ARG}^{83}$ and $\text{GLU}^{42}$ were observed among top-scored compounds.
The prevalence of these major pharmacophore interactions resulted in top-scored compounds exhibiting structure patterns akin to the majority of binding modes observed in the SOS1 case.
Conversely, the EGFR, JAK2, and PARP1 receptors predominantly comprised hydrophilic residues.  
In the case of the EGFR receptor, a solitary aromatic side chain, $\text{PHE}^{856}$, was found buried within the inner region of the binding pocket, while several residues formed hydrogen bonds with ligands, such as $\text{ASP}^{854}$ and $\text{ASP}^{855}$.
For the JAK2 receptor, all top-10 interactions were hydrogen bonds, with $\text{LEU}^{86}$ and $\text{ARG}^{134}$ featuring prominently among the interacting residues. 
These hydrogen bonds acted as anchoring interactions, leading us to conjecture that top-scored ligands exhibit similar docking poses. 
Similarly, in the case of PARP1, the majority of top-1000 ligands engaged in pi-pi stacking with $\text{TYR}^{246}$ and formed hydrogen bonds with $\text{ARG}^{217}$ and $\text{TYR}^{228}$. 
In the above three cases, the absence or scarcity of aromatic side chains resulted in top-scored compounds lacking structural pattern akin to those observed in SOS1 and ESR2. 
Instead, hydrogen bonds served as key interactions, with some functioning as anchoring interactions.

In Figure \ref{fig:ESR2_PGR}, it is evident that ESR2 and PGR primarily engage through hydrophobic interactions, with the exception of the PHE92 residue in PGR. These interactions play a pivotal role in constraining the ligand's 3D conformation, thereby influencing the accurate prediction of docking scores by 2D GNN models. Remarkably, as depicted in Figure \ref{fig:3d_similarity}, ESR2 exhibits considerable similarity in docking poses, whereas PGR does not. This disparity can be attributed to the conformational differences in binding sites of ESR2 and PGR. Specifically, while ESR2's binding site adopts a closed form, PGR's binding site remains open. The openness of PGR's binding site introduces variability in docking poses, thereby complicating the 2D GNN model's ability to predict docking scores accurately.
In contrast, the PGR receptor exhibited a scarcity of common pharmacophore interaction pattern, with pi-pi stacking involving $\text{PHE}^{92}$ being the sole prevalent pattern.
This finding aligns with the absence of similar binding pose pattern, as depicted in Figure \ref{fig:3d_similarity}.

In summary, as evidenced by the cases of SOS1, EGFR, ESR2, JAK2, and PARP1, critical features of the binding pocket, such as its pocket shape and side-chain population, affected the binding poses resulting from docking simulation and the occurrence of common pharmacophore interactions patterns.
During the active learning process, acquisition functions tended to select compounds whose structural patterns were associated with such 3D-shapes and pharmacophore interactions. 
By using top-scored compound results, as an acquisition step increased, the surrogate GNN model acquired an understanding of the relationship between compound structure and docking score.
Consequently, we conclude that receptors with well-defined binding pockets and interaction patterns constitute significant factors for the successful recovery of top-scored compounds in an active learning scenario.

\subsection{Surrogate models memorize simple structural patterns}
Our next investigation delves into whether surrogate GNN models learn the underlying principles governing ligand-protein interactions or simply memorize structural patterns found in top-scored ligands, as we mentioned earlier, such as shape and interaction similarities. 

\begin{table}[]
\caption{Molecular descriptors used in the linear dependency experiment.}
\label{tab:linear_regression}
\begin{tabular}{|l|l|l|}
\hline
\# of aromatic rings & \# of rotatable bonds              & \# of ketones \\ \hline
\# of rings          & \# of primary amines   & \# of esters  \\ \hline
\# of hydrogen bond donors           & \# of secondary amines & \# of amides  \\ \hline
\# of hydrogen bond acceptors           & \# of tertiary amines  & \# of ureas   \\ \hline
Molecular weight     & -                      & - \\ \hline
\end{tabular}
\end{table}

\begin{table}[]
\caption{Coefficient of determination ($R^2$) of linear regression fitting of docking score and prediction score with molecular descriptors.}
\label{tab:R2_linear_regression}
\begin{tabular}{|c|c|c|}
\hline
      & $R^2$ on docking score       & $R^2$ on prediction score     \\ \hline
SOS1  & 0.169                        & 0.576                         \\ \hline
EGFR  & 0.350                        & 0.684                         \\ \hline
ESR2  & 0.347                        & 0.522                         \\ \hline
JAK2  & 0.604                        & 0.691                         \\ \hline
PARP1 & 0.718                        & 0.752                        \\ \hline
PGR   & 0.102                        & 0.162                         \\ \hline
\end{tabular}
\end{table}

To explore the linear dependency between scores and simple molecular descriptors, two two linear regression models were trained using data points obtained at the last acquisition step. Each model predicted docking and GNN prediction scores using a set of molecular descriptors $\{ f_k \}$ outlined in Table \ref{tab:linear_regression}:
\begin{equation}
    \text{score} = \sum_{k} c_k f_k + \text{bias},
\end{equation}
where $c_k$ represents the linear coefficient for the $k$-th molecular descriptor. 
We chose these descriptors for linear regression based on several considerations: 
(1) aliphatic and aromatic rings typically serve as scaffolds, playing a pivotal role as key fragments,
(2) amine, amides, ureas, esters, and ketone groups are widely observed in fragment connections,
(3) the number of HBDs and HBAs is associated with hydrogen bonding, and
(4) the number of RBs correlates with the flexibility of the molecule's 3D-shape.

As depicted in Table \ref{tab:R2_linear_regression}, the coefficient of determination $R^2$ for the prediction scores exceeded those of the docking scores across all six receptors. This observation underscores the greater reliance of surrogate GNN models on specific structural patterns for fitting the docking score compared to AutoDock Vina.
Additionally, our analysis confirmed that linear fitting of both docking and prediction scores proved ineffective for the PGR case, potentially indicating a deficiency in similar structural patterns within acquired samples. 

\subsection{Screening power of the surrogate GNNs}
While surrogate GNN models exhibit a tendency to memorize common structural patterns rather than generalize underlying principles of protein-ligand interactions, 
active learning continues to wield significant influence in the discovery of top-docking-scored compounds, thereby substantially reducing computational resources, particularly in cases involving well-characterized binding pockets.
To assess the potential for further applications using surrogate GNN models, which may lack generalizability, we examined the screening efficacy of these models using the DUD-E dataset and 100M compounds in EnamineREAL library.

\begin{figure}[hbt!]
    \centering
    \includegraphics[width=0.95\textwidth,trim={0cm 0 0cm 0},clip]{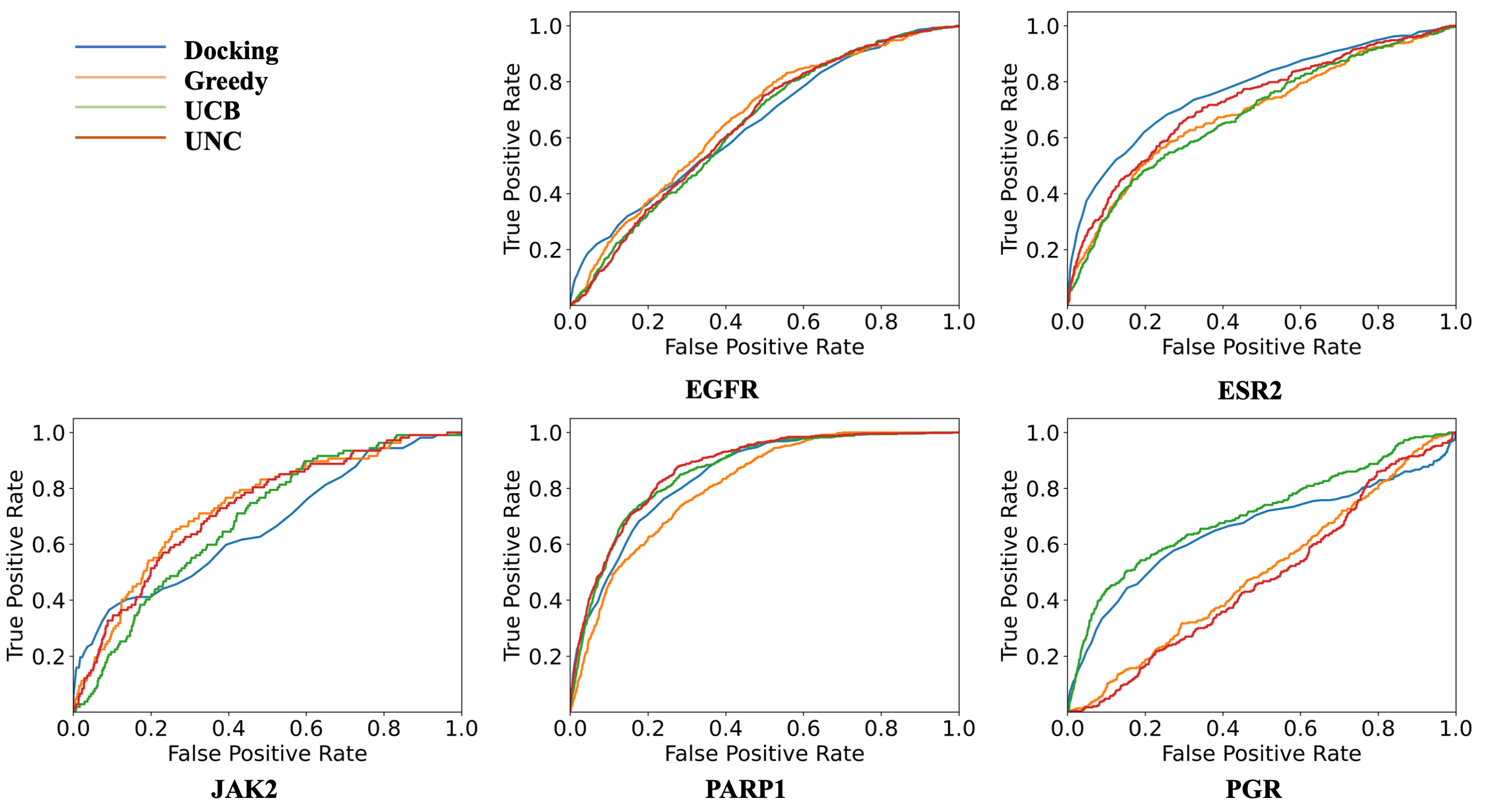}
    \caption{Receiving operation characeteristics (ROC) curves for screening with AutoDock Vina, Greedy, upper-confidence bound (UCB) uncertainty (UNC) acquisition strategies. Note that all the actives and decoys are from the DUD-E dataset.}
    \label{fig:ROC_curve}
\end{figure}

\begin{table}[hbt!]
\caption{Screening performance of Autodock-Vina and surrogate GNN models trained with Greedy, upper confidence bound (UCB), and uncertainty (UNC) acquisition strategies in terms of area under receiving operation curve (auROC). Note that all the actives and decoys are from the DUD-E dataset.}
\label{tab:auroc}
\begin{tabular}{|l|l|l|l|l|l|}
\hline
        & EGFR   & ESR2   & JAK2   & PARP1  & PGR    \\ \hline
Vina    & 0.649 & 0.772 & 0.662 & 0.845 & 0.652 \\ \hline
Greedy  & 0.665 & 0.690 & 0.735 & 0.809 & 0.498 \\ \hline
UCB     & 0.641 & 0.684 & 0.685 & 0.860 & 0.707 \\ \hline
UNC     & 0.645 & 0.725 & 0.725 & 0.869 & 0.476 \\ \hline
\end{tabular}
\end{table}
Initially, we applied Autodock-Vina (Vina) and the surrogate GNN models, acquired through Greedy, upper confidence bound (UCB), and uncertainty (UNC) acquisition strategies with the EnamineHTS library, to the DUD-E dataset. 
It is important to highlight that the surrogate GNN models were trained using compounds in the EnamineHTS library and were not exposed to the active and decoy compounds within the DUD-E set.
In Figure \ref{fig:ROC_curve} and Table \ref{tab:auroc}, we present the receiving operation characteristics (ROC) curves and area under ROC curves (auROC) for the five receptors within the DUDE-set.
While there is not a consistent superiority observed across all three surrogate GNN models, several acquisition instances demonstrate comparable or marginally improved performance compared to Vina.

\begin{figure}[hbt!]
    \centering
    \includegraphics[width=0.95\textwidth,trim={0cm 0 0cm 0},clip]{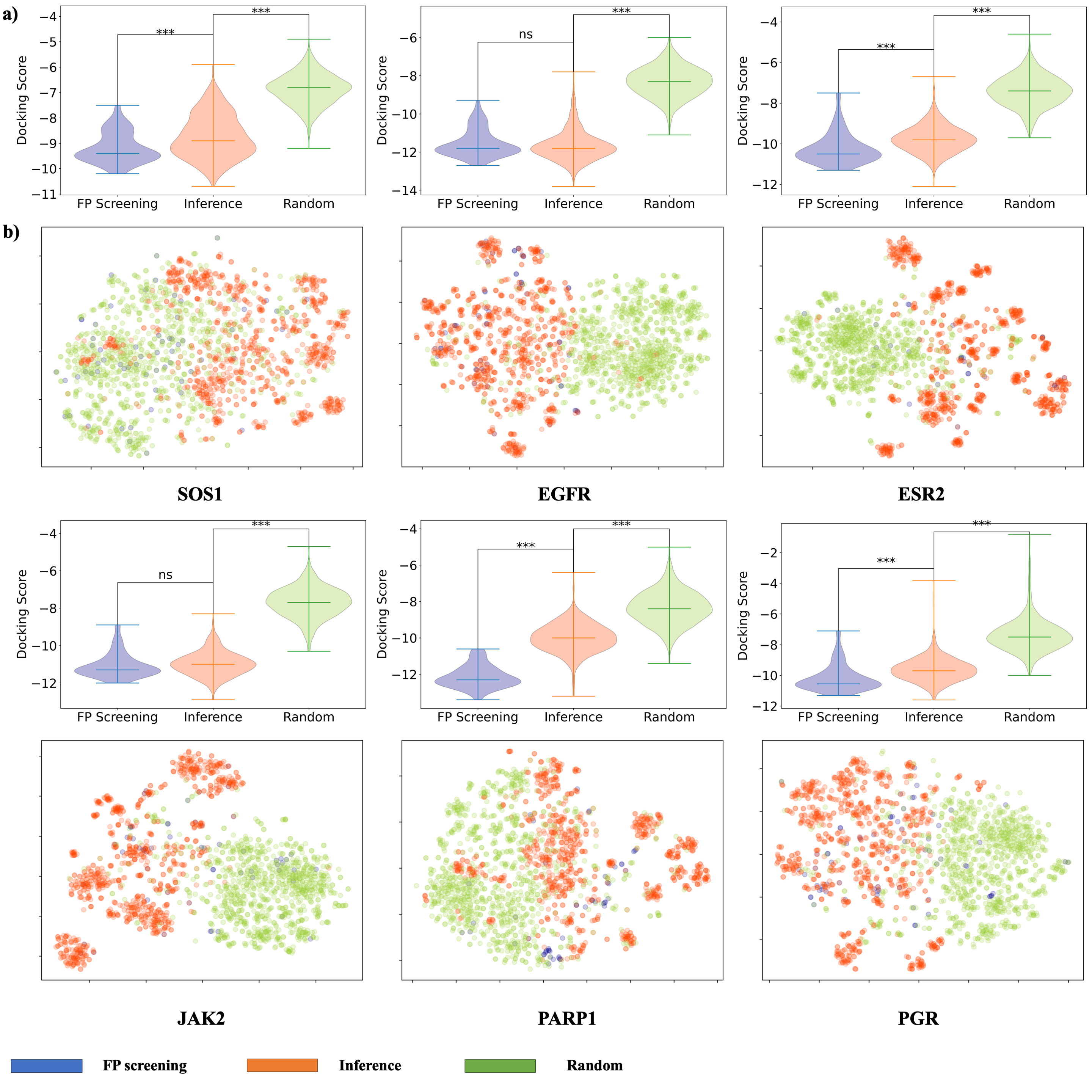}
    \caption{
    a) Distribution of docking scores for the three set of compounds. 
    b) Spatial distribution of the three set of compounds obtained by t-SNE plot of fingerprints generated using the RDKFingerprint function in RDKit. 
    FP-screening denotes a collection of compounds with a Tanimoto similarity greater than 0.85 with respect to the reference binder. 
    Inference refers to a set of compounds within top-1000 predictions. 
    Random stands for a selection of compounds sampled randomly from the EnamineREAL library.
    }
    \label{fig:matching}
\end{figure}


Despite the lack of consistent superiority of surrogate GNN models, their ability to predict docking scores significantly faster than Vina is noteworthy.
Therefore, we applied the GNN model obtained through UCB acquisition to screening 100M compounds randomly sampled from the EnamineREAL library, aiming to assess the effectiveness of GNN models on other larger datasets.
Illustrated in Figure \ref{fig:matching}, three distributions of docking score are visualized: (1) top-1000 predictions with the surrogate GNN, (2) randomly sampled 1000 compounds, and (3) compounds with a Tanimoto similarity of 0.85 or higher to the top 1000 docking-scored compounds from EnamineHTS library, with fingerprints obtained by using the RDKFingerprint function in RDKit.
Across all receptors, it was observed that the docking score distributions of top-1000 predictions exhibited a more downward shift compared to those of randomly sampled compounds. 
In two-dimensional chemical space visualization via t-distributed Stochastic Neighbor Embedding (t-SNE)\cite{van2008visualizing}, the spatial distribution of these two sets were distinctly separated.
Additionally, for the SOS1, EGFR, ESR2, and JAK2 receptors, compounds with docking scores higher than those screened by Tanimoto similarity were identified. 
Similarly, for the PARP1 and PGR receptors, compounds with docking scores comparable to those obtained through similarity-based screening were discovered using the surrogate GNNs, albeit the overall docking scores of the screened sets are higher for the similarity-based screening.

In summary, while the surrogate GNN models demonstrated screening performance comparable to AutoDock Vina, they did not consistently outperform it.
Nonetheless, leveraging these models allowed us to prioritize compounds with higher docking scores over randomly-sampled compounds within the broader chemical space of the EnamineREAL set, all while significantly reducing computational costs.
Moreover, the screening efficacy of the GNN models proved to be on par with or even superior to similarity-based screening.
Consequently, both two screening methods can be effectively employed to prioritize top-docking-scored compounds, which may help us to establish consensus on binding characteristics.
Particularly in a scenario lacking a reference binder, we advocate for the adoption of surrogate GNN models obtained with smaller yet diverse libraries as a good starting option for computationally efficient screening. Furthermore, for active learning pipelines in the EnamineREAL library, the surrogate model trained on the EnamineHTS library can serve as a valuable starting point.
 
\section{Conclusion}

While active learning of molecular docking typically does not integrate receptor information into the learning process of the structure-docking score relationship, previous works have illustrated its efficacy in recovering top-scored compounds with notable reductions in computational costs. 
In this work, we conduct a comprehensive benchmark of active learning of molecular docking across six receptors, aiming to elucidate the prediction behavior of surrogate models and shed light on how active learning selects top-scored samples and fits docking scores.
Our key findings are summarized as follows:
\begin{itemize}
    \item To maximize the discovery of top-docking-scored compounds, both Greedy and upper-confidence bound (UCB) acquisitions, which align with Bayesian optimization, are suitable strategies.
    \item Owing to biased sampling inherent in Greedy and UCB acquisitions, accurate prediction and rank-ordering are exclusively applicable to top-scored compounds.
    \item The success of active learning and the ensuing structural patterns observed in acquired compounds are significantly affected by the characteristics of the binding pocket, including pocket shape and potential interaction patterns. 
    \item Our analysis confirms a strong dependence of predicted docking scores on simple molecular structural descriptors, which led us to suspect that surrogate predictors tend to memorize the simple structural patterns rather than generalize principles in protein-ligand interactions.
    \item Despite surrogate models exhibiting performance akin to AutoDock Vina in distinguishing actives from decoys in the DUD-E dataset, their utilization proves effective in screening larger chemical spaces with reduced computational costs, as evidenced by our screening of the EnamineREAL library. 
\end{itemize}

While surrogate models may lack generalization, active learning remains a versatile tool with numerous applications. 
Firstly, it enables the discovery of top-scored compounds at reducedcomputational costs. 
Our validations have demonstrated the identification of common structural and interaction patterns.
Leveraging these patterns, active learning can identify and categorize regions crucial for tight-binding, facilitating the establishment ofstructure-activity relationships and the design of fragments for further fragment-based design.
Additionally, as illustrated, employing a surrogate model trained on a smaller yet diverse dataset can aid in screening a broader chemical space to prioritize top-scored compounds. 
Although not explored in this study, two-stage active learning is a viable approach, utilizing smaller and larger pools of compounds for the first and second stages, respectively.
In the initial stage, diverse ligand structures can be discovered, followed by a second stage focused on the search for more targeted fragments.

However, several limitations must be addressed to ensure more practical and reliable use of active learning.
As most machine learning models rely on the assumption of smoothness\cite{chapelle2009semi}, a surrogate model may not explain activity cliff in medicinal chemistry, where subtle changes molecular structure can lead to significant variations in molecular activity. 
Additionally, while a high `docking score' often suggests a high `binding affinity' in experimental assays, this correlation is not guaranteed, as highlighted by the lesson of \textit{the higher, the better},\cite{gorgulla2020open}.
Therefore, top-scored compounds acquired from active learning protocols should be evaluated using additional tools, such as visual inspection and free energy calculations, to screen out false positives. 
Considering receptor flexibility, active learning should ideally be conducted with various receptor conformations and consensus scoring methods.

Nevertheless, we believe that our research contributes to the understanding of how active learning operates in the drug discovery domain and prompts consideration of more effective utilization strategies. 
For instance, with the increasing comprehension of machine learning techniques and computational infrastructure, the iterative process of computational prediction and experimental validation, known as human-in-the-loop, has become prevalent in the computational discovery of materials and drug candidates.
As human-in-the-loop can be viewed as active learning with an experimental oracle, a thorough understanding of predictions made by surrogate models becomes crucial for fine-tuning of campaigns and candidate selections.
In summary, we anticipate that our analysis will enhance the efficiency and effectiveness of drug and material discovery using machine learning techniques.

\section{Methods}
\subsection{Docking simulation}
We used AutoDock Vina\cite{trott2010autodock} as a molecular docking tool.
A simulation center of each receptor target is extracted from the crystal ligand pose in the PDB complex structure. 
The options for docking simulations were as follows:
 \begin{lstlisting}
 --size_x = 20.0
 --size_y = 20.0
 --size_z = 20.0
 --exhaustiveness = 8
 --energy_range = 4
 --cpu = 1
 --num_modes = 1
 \end{lstlisting}

\subsection{Atom and bond descriptors}
We assigned the initial node descriptors (features) using the RDKit\cite{landrum2013rdkit} functions as follows:
\begin{itemize}
    \item One-hot encoding of atom types: atom.GetSymbol()
    \item One-hot encoding of atom degree: atom.GetDegree()
    \item One-hot encoding of number of hydrogens: atom.GetTotalNumHs()
    \item One-hot encoding of implicit valence number: atom.ImplicitValence()
    \item Indicator value for whether given atom is aromatic: atom.GetIsAromatic()
\end{itemize}
Also, the initial edge descriptors (features) were assigned as follows:
\begin{itemize}
    \item One-hot encoding of atom bond types: whether the bond type is single, double, triple or aromatic bond.
    \item Indicator value for whether bond is conjugated: bond.GetIsConjugated()
    \item Indicator value for whether bond is in ring: bond.IsInRing()
\end{itemize}
 
\subsection{Training surrogate GNN models}
For a surrogate GNN model\cite{battaglia2018relational}, we implemented the graph isomorphism network incorporating edge features (GIN-E) in node feature aggregations, which is a modified version of GIN in \citeauthor{xu2018powerful}\cite{xu2018powerful} and proposed in \citeauthor{hu2019strategies}\cite{hu2019strategies}.

Following the initialization of atom and bond descriptors, we preprocessed the input graph structure $G(X,E)$ of molecules, where $X$ represents the set of atom (node) feature vectors and $E$ represents the set of bond (edge) feature vectors.
Subsequently, for the node and edge feature sets $X = \{x_i\}$ and $E = \{e_{ij}\}$, where $x_i \in \mathbb{R}^{d_n}$ denotes the $i$-th node feature and $e_{ij} \in \mathbb{R}^{d_e}$ represents the edge feature between $i$-th and $j$-th node if two nodes are connected by a chemical bond, the first embedding layer linearly transformed the initial feature vectors to the pre-defined node and edge feature dimension:
\begin{equation}
\begin{split}
    & h_i^{(0)} = W_{n,emb} x_i \\
    & e_{ij}^{(0)} = W_{e,emb} e_{ij},
\end{split}
\end{equation}
where $W_{n,emb} \in \mathbb{R}^{d \times d_n}$ and $W_{e,emb} \in \mathbb{R}^{d \times d_e}$ are weight parameters, and $d_n$, $d_e$ and $d$ are node, edge and transformed feature dimensions.

At the $l$-th node update layer of GIN-E, for $l \in \{1, ..., L\}$, the node feature is updated from $h^{(l-1)} \in \mathrm{R}^{d}$ to $h^{(l)} \in \mathrm{R}^{d}$ as follows:
\begin{equation}
     h^{(l)}_{i} = \text{DO}(\text{LN}(h^{(l-1)}_{i} + \text{MLP}^{(l)}( \sum_{j \in \mathcal{N}_{i}} h_{j}^{(l-1)}+e_{ij}^{(l-1)}))),
    \label{eq:GINEupdate}
\end{equation}
where $\text{DO}$ and $\text{LN}$ stands for dropout\cite{srivastava2014dropout} and layer normalization\cite{ba2016layer}, and $\text{MLP}$ is two-layer multi-layer perceptron with non-linear activation $\sigma$:
\begin{equation}
    \text{MLP}^{(l)}(x) = W_2^{(l)}\sigma(W_1^{(l)}x + b_1^{(l)}) + b_2^{(l)}.
\end{equation}

To aggregate the updated node features, we utilize pooling by multi-head attention\cite{lee2019set, ryu2022accurate}, assigning different atom-wise importance (attention weight) to each node:
\begin{equation}
    z_{G} = \sum_{v} \alpha_v h_v^{(L)},
\end{equation}
where $\alpha_v$ is the attention weight assigned to the $v$-th node given by
\begin{equation}
    \alpha_v = \text{softmax}(\frac{1}{\sqrt{d}} \textbf{1} (W_{\text{pma}} h_{v}^{(L)})^{\top} ),
\end{equation}
where $W_{\text{pma}} \in \mathrm{R}^{d \times d}$ is a weight parameter and $\textbf{1} \in \mathrm{R}^{d}$ is a one-initalized seed vector of size $d$ and $W_{pma}$ is a weight parameter.

Finally, for the $n$-th sample in the training dataset, predictive mean $\hat{y_n}$ and uncertainty $\hat{y}_n$ of docking score are obtained by the linear transformation layer:
\begin{equation}
\begin{split}
    & \hat{y}_n = W_m^\top z_{G_n} + b_m \\
    & \hat{\sigma}_n = W_u^\top z_{G_n} + b_u
\end{split}
\end{equation}
where $W_m, W_u$ and $b_m, b_u$ are weight and bias parameters, respectively.

For gradient-descent minimization of learning objective, we implemented heteroscedastic mean-squared error loss\cite{nix1994estimating},
\begin{equation}
    L(\{y_n, \hat{y}_n, \hat{\sigma}_n\}) = \frac{1}{N} \sum_{n=1}^{N} \frac{1}{2\hat{\sigma}_{n}^{2}} (y_n - \hat{y}_n)^2 + \frac{1}{2} \log \hat{\sigma}_{n}^{2},
\end{equation}
to estimate both predictive docking score $\hat{y}_n$ and its uncertainty $\hat{\sigma}_n$.
Note that the predictive uncertainty $\hat{\sigma}_n$ is used in UCB and Unc acquisition methods. 

\subsection{Neural network implementation}
We used PyTorch\cite{paszke2017automatic} and Deep Graph Library\cite{wang2019deep} for implementation of GNN models. 
The hyperparameters and training configurations are detailed in the  table below.

\begin{table}[]
\begin{tabular}{|l|l|}
\hline
Number of node update layers ($L$) & 4          \\ \hline
Node feature dimension ($d$)       & 128        \\ \hline
Number of attention heads          & 4          \\ \hline
Number of training epoches         & 150        \\ \hline
Optimizer                          & AdamW\cite{kingma2014adam, loshchilov2017decoupled}      \\ \hline
Learning rate at 0/40/80/120 epoch & $10^{-3}$/$10^{-4}$/$10^{-5}$/$10^{-6}$   \\ \hline
Dropout probability                & 0.2        \\ \hline
Weight deacy                       & $10^{-6}$  \\ \hline
\end{tabular}
\end{table}

\section{3D pose distance and multi-dimensional scaling}
To assess the 3D-conformational similarity between two given molecules, we computed the 3D pose distance based on their docking poses.
Utilizing RDKit, we extracted the 3D conformations of the docking poses (\texttt{Chem.MolFromMol2File} function) and subsequently calculated the 3D-shape Tanimoto distance (\texttt{rdShapeHelpers.ShapeTanimotoDist} function) between two given poses. 
Then, we built the distance matrix where $(i, j)$-th element is given by 
\begin{equation}
    \text{Dist}_{ij} = \dfrac{1}{(\text{3DSim$_{ij}$} + \epsilon) \sqrt{MW_i MW_j}},
\end{equation}
where $MW_i$ is the $i$-th sample's molecular weight and $\epsilon=0.01$ were employed to avoid divied-by-zero error.
Note that we divided this distance into molecular weight for scaling with molecular size. 
Using the distance matrix, we utilized multidimensional scaling (MDS)\cite{cox2008multidimensional}, implemented in scikit-learn \cite{kramer2016scikit}, for two-dimensional visualization of conformational space. 

\section*{Code availability}
Our implementation and scripts for data analysis are available at \url{https://github.com/jasonkim8652/al_breakdown}.

\section*{Acknowledgements}
We acknowledge Juwon Hong for proofreading the manuscripts.
This research was supported by Galux Inc. (No. Galux-20210001) and the National Research Foundation (NRF) funded by the Korean government (MSIT) (RS-2023-00232157). 

\section*{Author contributions}
Idea conception, implementation, and experiments were undertaken by Seongok Ryu and Jeonghyeon Kim. 
All three authors collectively analyzed the results and contributed to the writing of the manuscript.
\bibliography{achemso-demo}
\end{document}